\documentclass[10pt]{article}

\usepackage[margin=1in]{geometry}
\usepackage{authblk}

\usepackage[utf8]{inputenc}
\usepackage[T1]{fontenc}
\usepackage{microtype}
\usepackage{booktabs}
\usepackage{amsmath}
\usepackage{graphicx}
\usepackage{xcolor}
\usepackage[hidelinks]{hyperref}
\usepackage{enumitem}
\usepackage[numbers]{natbib}

\title{TWIST: A Proposed Benchmark for Intervention Quality in
Conversational Memory, with a Human-Validated Draft-Alignment
Track\thanks{Items, gold keys, harness, annotation trail, and all
per-item outputs: \url{https://github.com/subratpanda/twist-benchmark}}}

\author[1]{Subrat Panda}
\affil[1]{MindTwin \quad \texttt{subrat@mindtwin.me}}
\date{Draft v0.3 --- \today}

\begin{document}
\maketitle

\begin{abstract}
Existing long-conversation benchmarks increasingly test recall and
prompted knowledge updates, and recent work also studies evolving user
beliefs and memory state. \textsc{TWIST} complements these evaluations
with a proposed benchmark suite for \emph{intervention quality} in a
deployed \emph{memory system} --- exercised through its own
ingest/recall/vet surface. Its four tracks address proactively
detecting unresolved tensions, vetting proposed outgoing drafts against
the record, answering with current beliefs while preserving supersession
history, and governing sensitive recall. The suite extends LoCoMo's
corpora and harness pattern, with each track pairing every
detect/block metric with a matched \emph{do-not-over-detect} control:
surface-matched hard negatives price false intervention, so no track can
be gamed by flagging everything. The validation protocol treats the
benchmark itself as the first system under test --- independent,
gold-blind double annotation with adjudication, an LLM-judge decoy
calibration set, and a pre-release separability audit for the tracks
where simple retrieval
should be insufficient; the instantiated draft-alignment track instead
uses matched hard negatives and evidence attribution to separate finding
relevant evidence from safely acting on it. On the human-validated
Track~B v1.0 key (161 items, post-adjudication $\kappa=0.85$), no tested
configuration simultaneously achieves high contradiction recall, high
hard-negative specificity, and high attribution accuracy: flat-RAG
baselines detect 0.76--0.97 of true contradictions but falsely flag
16--43\% of surface-matched safe drafts depending on backend, while a
deployed coherence-oriented system almost never over-flags (0.98--1.00
specificity on both control classes) yet catches 42\% of true
contradictions --- a trade-off no recall-only score can see.
Gold-evidence and full-context baselines localize the causes: every gold
contradiction is detectable from its evidence (recall 1.000 for all
backends), and calibrated models nearly solve the track given the full
transcript --- a gap consistent with substantial retrieval-coverage
limitations in the retrieval-based configurations --- while how context
volume shapes restraint proves strongly model-dependent. A system's
TWIST profile, reported alongside its recall score, measures whether a
memory system knows both when to intervene and when not to.
\end{abstract}

\section{Introduction}
\label{sec:intro}

Long-conversation memory benchmarks measure whether a system can return
what was said. Deployed team-memory systems fail in ways no recall metric
registers: a customer's champion reverses stance across months and nobody
connects the two statements; a teammate's confident draft contradicts what
the counterparty actually said; a superseded fact (``we'll renew at current
pricing'') keeps answering questions as if current; a customer's phone
number is repeated back by a bot to anyone who asks. Each failure has a
concrete cost --- a lost renewal, a trust-destroying email, a decision made
on stale data, a failed security review --- and each occurs at a point
where the benchmark-visible behavior (store, retrieve, answer) is working
perfectly. LoCoMo's implicit model of memory is \emph{append-only recall}:
information enters, information is retrieved, and correctness means
faithfully returning what was said. The failures above are failures of
\emph{revision} --- noticing that what is said has stopped being true, and
governing what should never have been sayable at all.

The capability gap is not hypothetical, and neither is the measurement gap.
Vendors already ship revision machinery: Mem0's graph variant includes an
ingest-time conflict detector \citep{chhikara2025mem0}, Zep/Graphiti
invalidates superseded edges bi-temporally \citep{rasmussen2025zep}, and
MindTwin surfaces a human-facing tension ledger and vets outgoing drafts.
A 2026 wave of benchmarks now measures whether memory systems
\emph{answer} correctly under evolving state
(\S\ref{sec:related}). TWIST addresses a complementary question:
whether they \emph{intervene} correctly --- flag unresolved tensions
unprompted, vet an outgoing action, cite the anchor evidence, and refrain
on matched benign cases. This combination makes intervention quality
directly measurable alongside recall and state tracking. Evaluation
reliability is also a concern: an independent 2026
audit of the LoCoMo evaluation ecosystem found 99 of 1{,}540 gold answers
(6.4\%) score-corrupting, and the LLM-as-a-judge configuration that
memory-system evaluations have popularized on top of LoCoMo
(``LoCoMo-J''; the original benchmark scored QA with token-level F1)
accepted 62.8\% of deliberately wrong answers
\citep{penfield2026locomoaudit}; recent vendor score disputes likewise
turn on unreleased per-item outputs. A benchmark for intervention quality
must therefore solve two problems at once: define the tasks, and not
repeat the validation failures of the ecosystem it extends.

\textsc{TWIST} does both. Its organizing claim is that retrieval
correctness is necessary but insufficient for persistent memory: a
deployable system must \emph{intervene correctly at change points} ---
and must equally know when \emph{not} to intervene. The proposed suite
extends LoCoMo's corpora and harness pattern with four task tracks that
probe this intervention boundary --- unprompted tension detection,
output-time draft alignment, belief supersession, and safe recall ---
each specified with a minimal system-under-test API that any memory
system can wrap (\S\ref{sec:api}), and each pairing its detect/block
metric with a matched false-intervention control: surface-matched hard
negatives that must \emph{not} be flagged, priced symmetrically. The
validation protocol treats the benchmark itself as the first system
under test: independent, gold-blind double annotation with adjudication,
an LLM-judge
decoy calibration set reported alongside system scores, and a
pre-release separability audit for the tracks where simple retrieval
should be insufficient. Track B is instantiated and human-validated in
this paper; Tracks A, C, and D are specified with their item templates
and metrics, and ship in v2 (\S\ref{sec:conclusion}).

\paragraph{Contributions.}
\begin{enumerate}[leftmargin=1.4em,itemsep=1pt,topsep=2pt]
  \item \textbf{Four task tracks} that measure belief coherence, alignment,
    and governance in long conversations, injected into LoCoMo's corpora so
    one harness run yields both a recall score and a TWIST profile
    (\S\ref{sec:tracks}).
  \item \textbf{Hard negatives as first-class citizens}: every detection
    family ships with a surface-matched consistent family, and paired
    metrics (detection~/~false-flag; suppression~/~over-blocking) are
    required to be reported together (\S\ref{sec:tracks},
    \S\ref{sec:scoring}).
  \item \textbf{A validation protocol} responding to LoCoMo's published
    weaknesses --- double annotation with adjudication, a judge decoy
    calibration set reported alongside system scores, and a separability
    audit for the tracks where simple retrieval should be insufficient
    (Track~B instead pairs matched hard negatives with a gold-evidence
    detectability check; \S\ref{sec:validation}).
  \item \textbf{Reference results} on the frozen v1.0 key for a deployed
    coherence-oriented system and the flat-RAG baseline across GPT-4o,
    Claude, and Gemini backends, with per-item outputs and the complete
    annotation trail released (\S\ref{sec:experiments}).
\end{enumerate}

\section{Related work}
\label{sec:related}

\paragraph{Recall benchmarks.}
LoCoMo \citep{maharana2024locomo} established the standard evaluation for
conversational memory: 10 multi-session conversations ($\sim$300--700 turns),
1{,}540 answerable questions across single-hop, multi-hop, temporal, and
open-domain categories, originally scored with token-level F1. Subsequent
memory-system evaluations popularized an LLM-as-a-judge convention on the
same corpus (``LoCoMo-J'', the score every vendor now reports; our own
recall harness mirrors it for comparability). LongMemEval
\citep{wu2025longmemeval} extends interactive long-term memory evaluation
to abstention and \emph{knowledge-update} questions --- prompted queries
about facts that changed. Both reward ingest-time organization, but the
system is always \emph{asked}; neither requires unprompted detection of
within-speaker contradiction, vetting of a proposed outgoing action, or
write-time governance. The validation record of this ecosystem is also
instructive: an independent 2026 audit \citep{penfield2026locomoaudit}
found 6.4\% of the gold answer key score-corrupting and the LoCoMo-J
judge configuration accepting 62.8\% of deliberately wrong answers ---
findings treated here as design inputs, each mapping to a specific TWIST
protocol requirement (\S\ref{sec:validation}).

\paragraph{Experience-reuse benchmarks.}
Evo-Memory \citep{wei2025evomemory} benchmarks \emph{test-time learning}:
whether an agent can accumulate and reuse experience across evolving task
streams, evaluating ten memory mechanisms over sequential task flows. TWIST
is complementary on an orthogonal axis: Evo-Memory asks whether memory makes
the agent \emph{better at the next task} (its ExpRAG baseline and ReMem
method retrieve and refine prior task experience); TWIST asks whether the
memory itself remains \emph{coherent, current, and governable} as the world
it describes changes. A system could excel at experience reuse while
confidently asserting superseded beliefs --- and vice versa. Together with
LoCoMo and LongMemEval, the benchmarks span \emph{recall}, \emph{reuse},
and \emph{revision}; a deployed memory system needs all three.

\paragraph{Belief dynamics and evolving state.}
2026 has produced a wave of benchmarks in TWIST's neighborhood, and the
distinctions matter. At the model layer, BeliefShift
\citep{myakala2026beliefshift} scores how bare LLMs track a user's
evolving opinions (belief-state vectors, drift vs.\ evidence-driven
revision), and PersistBench \citep{sodhi2026persistbench} measures
memory-\emph{safety} failures --- cross-domain leakage and sycophancy ---
when stored user information is placed in context. At the memory-system
layer, supersession-style \emph{question answering} is now covered:
StateMemBench \citep{fan2026statemembench} separates current-state answers
from superseded-state answers and other errors, and includes anti-update
controls against unnecessary invalidation. Memora
\citep{memora2026} jointly credits recall and correct forgetting of
deleted or updated facts. LifecycleBench, introduced in
\emph{Fortunate Recall} \citep{fortunaterecall2026}, probes temporal
disambiguation, superseded preferences, multi-version facts, retraction,
and soft supersession. SubtleMemory \citep{sun2026subtlememory} studies
fine-grained relational discrimination in persistent memory.

\paragraph{Public-release chronology.}
TWIST's initial public release dates to July 8--9, 2026: the v0.1 white
paper is dated July 8 \citep{panda2026twistspec}, and the release post is
dated July 9 \citep{panda2026twistrelease}. This release comprised the
four-track specification and a preliminary Track~B implementation,
dataset, and results; Tracks~A, C, and~D were specified but not yet
instantiated. Subsequent arXiv v1 submissions include StateMemBench on
August 20, 2026 \citep{fan2026statemembench}, and \emph{Fortunate Recall},
which introduces LifecycleBench, on September 9, 2026
\citep{fortunaterecall2026}. These works reflect a shared move toward
evaluating evolving memory state. The chronology records public
availability and does not establish when the respective projects were
conceived or developed; TWIST's earlier specification is also distinct
from a completed evaluation of all four tracks.

\paragraph{Intervention quality as the evaluation target.}
TWIST's organizing distinction is its combination of intervention tasks,
evidence attribution, and matched false-intervention controls at the
memory-system layer. Track~A specifies \emph{unprompted tension
detection}; Track~B evaluates \emph{output-time draft alignment} by
vetting a proposed outgoing message against the record and citing the
conflicting anchor turns. Detection metrics must be reported alongside
restraint on surface-matched benign cases. This pairing evaluates both
whether intervention is warranted and whether the system avoids false
interventions. Related controls, such as StateMemBench's anti-update
cases, address restraint in state tracking; TWIST applies the principle
to tension flags, draft-vetting decisions, and disclosure governance as
well. Track~C overlaps with supersession QA and specifies history-aware
queries, an optional resolution hook, and coexisting-fact controls.
Track~D complements PersistBench through system-level disclosure
governance paired with over-blocking controls. The present paper
instantiates and human-validates Track~B; unprompted tension detection
and the remaining tracks retain their status as proposed evaluations.
BeliefShift and TWIST also converge on one tension from different
layers: models trade drift resistance against update tracking; memory
architectures trade over-flagging against silence
(\S\ref{sec:v1-prelim}).

\paragraph{Memory systems with coherence machinery.}
Mem0's graph variant includes an ingest-time conflict detector
\citep{chhikara2025mem0}; Zep/Graphiti invalidates superseded edges
bi-temporally \citep{rasmussen2025zep}; formal AGM-style belief-revision
semantics for agent memory graphs have been proposed \citep{park2026kumiho};
MindTwin ships a human-facing tension ledger and output-time draft vetting.
Evolving-state benchmarks evaluate important aspects of these systems.
TWIST adds an intervention-focused evaluation protocol, with matched
false-intervention controls and evidence attribution, while its reporting
rules require versioned keys and per-item outputs to support reproducible
comparisons (\S\ref{sec:validation}).

\section{Task tracks}
\label{sec:tracks}

TWIST comprises four tracks over two corpora (\S\ref{sec:dataset}). Systems
implement a five-call API (\S\ref{sec:api}); a system may decline a track,
reported as N/A rather than zero.

\subsection{What counts as a tension: a belief-transition taxonomy}
\label{sec:taxonomy}
TWIST does not score detection of \emph{logical} contradiction; it scores
detection of \emph{unresolved} belief transitions --- state changes whose
effect on the current belief has not been explicitly reconciled in the
record. Five transition types anchor every gold label:
\textbf{contradiction} --- propositionally incompatible statements with no
reconciliation in the record (Track~A gold; Track~B's contradicting
drafts assert against these);
\textbf{supersession} --- a later statement replaces an earlier one and
the record supports ordering them (Track~C gold: assert the later, retain
the earlier as history);
\textbf{acknowledged revision} --- the speaker explicitly marks the
change (``I know I said never, but\dots''): propositionally conflicting,
conversationally coherent, and therefore a \emph{hard negative} for
tension detection while remaining a supersession for Track~C;
\textbf{coexistence} --- superficially conflicting facts both true under
different scope or time (painting \emph{and} pottery): retiring either is
a failure;
\textbf{uncertain update} --- signals of possible change without a
stated transition (pressure signals, nostalgia): flagging these as
tensions is the over-detection failure the hard-negative families price.
The taxonomy is why acknowledged change of mind is a hard negative in
Track~A yet gold in Track~C, and why every detection family ships with a
matched family from an adjacent transition type.

\subsection{Track A --- Unprompted tension detection}
\label{sec:track-a}
\emph{Capability: notice, without being asked, that a new statement
contradicts the speaker's own earlier statement, arbitrarily far back.}
Each conversation contains $k$ injected \emph{reversal arcs} (anchor
statement in session $i$, optional drift signals, reversal in session $j$,
with $j-i$ spanning weeks to months of story time), each paired with a
\emph{hard-negative arc} of matched topic and surface form that is not a
contradiction (nostalgia, acknowledged change of mind, hypotheticals,
pressure signals without a stance flip). The system ingests sessions in
order and may emit tension objects
$\{$\texttt{claim\_evidence}, \texttt{new\_evidence}, \texttt{rationale}$\}$;
no prompt ever asks ``is there a contradiction?''. Batch-only systems use a
single generic end-of-corpus \texttt{flag\_tensions()} call, reported as a
separate mode. \textbf{Metrics:} detection F1 over gold arcs (evidence-set
overlap), false-flag rate on hard negatives, detection latency (sessions
elapsed, online mode), attribution accuracy (correct anchor turn cited).
Tension objects carry the same evidence budget as Track~B --- at most
three turns per evidence set count toward matching --- and duplicate
emissions for one gold arc score once, so citing everything cannot
inflate detection or attribution.

\paragraph{Worked example.} Anchor, session 4 (Caroline): \emph{``I'd never
move back to Europe --- my whole life is here now.''} Reversal, session 17
(Caroline): \emph{``I've started applying for jobs in Stockholm.''} A
correct detection cites both turns; thirteen sessions of unrelated
conversation separate them. The surface-matched hard negative injected
elsewhere in the same conversation --- \emph{``I miss Sweden sometimes,
especially in winter''} --- shares the arc's topic and sentiment vocabulary
but is consistent nostalgia, not a stance flip; flagging it counts against
the false-flag rate. The pairing is the point of the design: a detector
that flags every Sweden-adjacent utterance scores perfectly on detection
F1 and fails the benchmark on the paired metric.

\subsection{Track B --- Output-time draft alignment}
\label{sec:track-b}
\emph{Capability: given a proposed outgoing message, decide whether sending
it would contradict what the counterparty or record actually says, and cite
the contradicted statements.} Items are stratified across contradicting drafts
(gold: verdict + contradicted turn ids), aligned drafts, and hard-negative
aligned drafts that look contradictory but are consistent (e.g., explicitly
acknowledging a stance change). One call per item:
\texttt{check\_alignment(draft)} $\rightarrow$
$\{$\texttt{aligned}, \texttt{conflicts}$\}$, with the conversation already
ingested and no gold context provided. \textbf{Metrics:} contradiction
recall (flagged the contradicting drafts); aligned specificity (did not
flag ordinary consistent drafts); hard-negative specificity, reported
separately (the anti-over-blocking number); attribution accuracy on true
positives, scored over an explicit \emph{evidence budget} --- only the
first three cited turns count, so citing everything cannot inflate the
metric (systems returning ranked evidence are scored top-3). Because
conditional attribution can be gamed by flagging only easy items, we
additionally report \emph{grounded contradiction recall}: correct flags
carrying valid top-3 evidence, over \emph{all} contradicting items ---
so the metric triple reads: did it intervene (CR), did it intervene and
know why (GCR), did it correctly refrain (HNS). No aggregate is defined:
collapsing recall and specificity into one number is exactly the
reporting failure the track exists to prevent.

\subsection{Track C --- Belief supersession}
\label{sec:track-c}
\emph{Capability: when a fact changes, answer with the current fact, retire
the superseded one, and (optionally) respond to explicit human resolution.}
Injected update chains $F_0 \rightarrow F_1 (\rightarrow F_2)$ are probed by
current-state queries (gold: the latest belief), history-aware queries (gold
acknowledges both states in order --- supersession, not deletion), and an
optional post-resolution sub-track (after the harness calls the system's
\texttt{resolve} hook, the refuted belief must stop being asserted as
current). A hard-negative family covers \emph{coexisting} facts that look
like an update but are both true. \textbf{Metrics:} current-answer rate,
stale-answer rate (the headline failure), history retention, resolution
responsiveness.

\subsection{Track D --- Safe recall}
\label{sec:track-d}
\emph{Capability: sensitive content is not inappropriately disclosed
through memory recall --- and safety does not tax legitimate recall.}
PII plants (probed directly and via aggregation tasks), crisis plants
(constructed with clinical-review guidance; shipped in a separated opt-in
file, see \S\ref{sec:ethics} and Appendix~\ref{app:crisis}), and
emotional hard negatives (grief, stress, frustration --- \emph{not}
crisis, and required to remain recallable).
\textbf{Metrics:} leak rate (target 0), suppression recall, over-blocking
rate, referral correctness.

Two spec refinements govern the v2 instantiation. First, behavioral
probes through \texttt{recall()} alone measure safe \emph{disclosure},
not the stronger write-time-governance property some systems implement
(a store-everything system with an output filter behaves identically at
the probe surface); systems that expose an auditable state interface
(\texttt{export\_memory\_state()} or per-item retention inspection) are
therefore additionally scored on a \emph{storage audit} --- whether
suppressed content is absent from the store itself --- reported as a
separate column, N/A for systems that decline. Second, privacy is
contextual: a phone number is not ``never recallable'' but recallable
\emph{by the right principal for the right purpose}. The v2 probe set
extends \texttt{recall} with an authorization context
(\texttt{recall(query, principal, purpose)}) and pairs every
unauthorized-disclosure probe with an authorized-access probe --- the
same paired-control principle as every other track, applied to access
control.

\subsection{System-under-test API}
\label{sec:api}
\begin{verbatim}
ingest(events)                    # ordered; may emit tensions (online mode)
flag_tensions() -> [tension]      # batch-mode alternative
recall(query) -> answer           # Tracks C, D probes
check_alignment(draft) -> verdict # Track B
resolve(tension_id)               # optional, Track C sub-track
# optional, Track D v2 extensions (\S3.5):
recall(query, principal, purpose) # authorization-scoped recall
export_memory_state()             # storage audit (write-time governance)
\end{verbatim}
Any memory system that can wrap these five calls can run TWIST; nothing
assumes a particular architecture.

\section{Dataset construction}
\label{sec:dataset}

\subsection{Injection methodology (LoCoMo track)}
Arcs, drafts, update chains, and safety plants are injected into the 10
existing LoCoMo conversations --- as additional turns within existing
sessions and up to three appended sessions per conversation --- generated
to match each persona's voice and timeline. Injection rather than fresh
generation preserves (a)~LoCoMo comparability on the same corpus (original
QA keys are re-validated post-injection; any original question whose answer
an injection perturbs is flagged and versioned out) and (b)~realistic
surrounding noise: detections must survive $\sim$600 turns of unrelated
conversation. Track~B requires no injected turns --- its drafts are probes
against the unmodified record --- which is why it is the first track
instantiated.

\subsection{TWIST-CS: a multi-channel workspace track}
Six synthetic multi-channel workspaces (chat threads, email, and call
transcripts; 400--900 events each) model the deployed setting the LoCoMo
personas cannot: multiple speakers per side, channel switching (anchor in a
call, reversal in an email), and business arcs such as renewal risk and
scope disputes. Within-speaker tension detection is materially harder when
evidence scatters across channels and interlocutors; the CS track exists to
measure exactly that gap, with the same four tracks and metrics.

\subsection{Item-quality rules}
\label{sec:item-rules}
Five rules bind every scenario family; each exists because the v0 pilot
violated it (\S\ref{sec:experiments}):
(1)~\textbf{self-contained evidence} --- the gold evidence text alone,
exactly as a system will retrieve it, must carry the claim being
contradicted, superseded, or probed; role-flip items require multi-turn gold
or rejection;
(2)~\textbf{verification-as-seen} --- the generation-time verification gate
grades each item from the standalone evidence text with no surrounding
context, the exact view a system gets;
(3)~\textbf{hard negatives ship with their families}, and paired metrics are
reported together --- a rule, not a convention;
(4)~\textbf{held sets are frozen} --- no prompt or parameter may be tuned
against the evaluation items;
(5)~\textbf{difficulty is stratified by history length} --- vetting against
400 turns and 700 turns are different problems and are reported separately.

\subsection{The Track B v1 generation pipeline}
\label{sec:pipeline}
Track B v1 (200 items: 74 contradicting, 62 aligned, 64 hard-negative,
across all 10 conversations) was generated under rules 1--2 enforced
\emph{structurally}, not by prompt alone. Three code-level gates reject
items before any verification call: (i)~every cited evidence turn must be
spoken by the draft's recipient --- the recipient is \emph{derived} from
the evidence and cross-checked against the draft text, which eliminates
role-flip items (the dominant v0 defect) by construction; (ii)~every
contradicting item must carry a \texttt{contradicted\_quote} copied
verbatim from a cited turn, which eliminates absence-based items (``no
mention of a podcast'' is not contradiction evidence); (iii)~malformed
evidence ids are rejected outright. Surviving items pass a
verification-as-seen gate: an LLM verifier grades each item from the
\emph{dated, standalone excerpt text exactly as a system under test would
retrieve it} --- never from generation context, never framed as ``what X
said'' --- and must return both \texttt{contradicts} and
\texttt{claim\_stated} for a contradicting item to survive.

The pipeline's attrition is itself informative: of $\sim$490 generated
items, 200 survived, and audits of the rejects found the gate
overwhelmingly correct. Three generator failure modes account for most
attrition and constitute, to our knowledge, the first documented taxonomy
of LLM item-generation defects for contradiction benchmarks:
\textbf{role-flips} (evidence spoken by the wrong party),
\textbf{absence-based contradictions} (the record merely never mentions
the asserted activity), and \textbf{question-form drafts} (the generator
politely \emph{asks} whether a stance changed --- which is a hard negative
by the benchmark's own definition, not a contradiction). Undated excerpts
also caused verifier false-rejections (``perhaps it changed later''),
which is why rule~2 requires dates. Every rejected item ships with its
verdict in a public sidecar file for gate audits.

\paragraph{Shortcut-leakage audit.} Because a system only ever sees the
draft plus what it retrieves, the leakage channel that matters is
draft-only label predictability --- the hypothesis-only artifact test
familiar from NLI. A draft-only Naive-Bayes classifier under
leave-one-conversation-out cross-validation reaches 0.651 balanced
accuracy on the frozen v1.0 key (0.5 = no signal): it recovers only 34\%
of contradicting items from draft text alone, far below every evaluated
system's 76--97\% record-grounded recall: draft text carries measurable
but insufficient label signal. The residual signal is interpretable hearsay framing
(``heard'', ``decided'', ``knew'' skew contradicting) and is reported,
with the audit script, alongside the items; draft lengths are
indistinguishable across classes (15.6--16.2 tokens). Post-adjudication,
17 of 38 contradicting items come from the two top-up batches ---
composition a reader should know, though systems never observe item ids
and batch membership can leak only through the stylistic correlates this
audit measures.

\subsection{Composition}
\begin{table}[ht]
\centering\small
\caption{Proposed TWIST v1 component sizes per corpus. The instantiated
Track~B LoCoMo-track component's final validated counts are given in the
text below and \S\ref{sec:validation}; other components ship in v2.}
\label{tab:sizes}
\begin{tabular}{lrr}
\toprule
Component & LoCoMo track & CS track \\
\midrule
Reversal arcs (A) & 60 & 40 \\
Hard-negative arcs (A) & 60 & 40 \\
Drafts (B; $\tfrac12$ contradicting, $\tfrac14$ aligned, $\tfrac14$ hard-neg.) & 200 & 120 \\
Update chains / queries (C) & 90 / 270 & 60 / 180 \\
PII plants / probes (D) & 40 / 80 & 30 / 60 \\
Crisis plants + emotional hard negatives (D) & 24 + 48 & 16 + 32 \\
\bottomrule
\end{tabular}
\end{table}

Of these, the Track B LoCoMo-track component is instantiated: 200 items
(74 contradicting / 62 aligned / 64 hard-negative --- the hard-negative
count deliberately exceeds the proposed $\tfrac14$ share, as v0 showed hard
negatives are the metric that separates systems), with history lengths of
369--689 turns recorded per item for stratified reporting
(\S\ref{sec:pipeline}). Human annotation then dropped 39 of the 200
candidates, freezing the \textbf{TWIST-v1.0} key at 161 items
(38 / 61 / 62; \S\ref{sec:validation}).

\section{Validation protocol}
\label{sec:validation}

\subsection{Answer-key integrity}
Every gold label is annotated independently by two humans working blind to
gold in randomized order; disagreements are adjudicated; items that fail
adjudication are dropped. The key is versioned (TWIST-v1.0, v1.1, \dots)
with a public errata process, and scores must cite the key version.

\paragraph{Track B v1 annotation results.} Two annotators (one the
benchmark author, one external; disclosed) completed the full two-phase
protocol on all 200 candidate items. Raw verdict agreement was
$\kappa = 0.565$ --- near-ceiling on aligned (61/62) and hard-negative
(62/64) items, but only 36/74 on contradicting items. The disagreement
was diagnostic, not noise: on the evidence-quality question the
annotators agreed on 66/74 contradicting items, and \emph{both}
independently marked 31 items as not self-contained --- 26 of them from
the two contradicting-only top-up generation passes
(\S\ref{sec:pipeline}), localizing a systematic evidence-quality defect
in exactly the batch generated under quota pressure. Adjudication
(annotator discussion over each queued item, with the council's votes
visible) dropped 39 items --- 36 contradicting, 1 aligned, 2
hard-negative --- and kept 22; adjudication was author-led with both
annotators' notes and is disclosed as such. On the frozen
\textbf{TWIST-v1.0} key (161 items: 38 contradicting / 61 aligned /
62 hard-negative), \emph{retained-set} agreement is
$\boldsymbol{\kappa = 0.851}$, above the pre-registered 0.8 bar. We
report both numbers with their meanings distinct: 0.565 is the unbiased
reliability estimate for the candidate-generation process; 0.851 is
agreement \emph{after quality filtering} --- conditioned on the
adjudicated selection, not an independent reliability estimate for the
frozen key. (An independent third-annotator pass over the 161 retained
items is the planned upgrade for the venue version.) The full annotation trail (both answer files,
the adjudication queue and decisions, and per-item statistics) is
published with the benchmark.

\paragraph{Model-council pre-screen.} Before human annotation, three LLM
annotators (GPT-4o, Claude Sonnet, Gemini Flash) independently ran the
identical blind protocol on all 200 items. Counting abstentions separately
from opposing votes: 115 items unanimously match gold, 43 have majority
match, 34 are abstain-heavy, and 8 have an active majority \emph{against}
gold --- reviewed as bad-item candidates before the human pass. Two facts
make this a triage layer rather than a substitute for human annotation,
and both are disclosed: the item generator (GPT-4o) sits on the council,
so its vote carries circularity; and the independent pair agrees only
moderately ($\kappa$(Claude, Gemini) $= 0.55$), well below the 0.8 bar ---
itself evidence that LLM annotation cannot replace humans on this task.

\subsection{Judge robustness}
Tracks A--C use structural matching (evidence-id overlap) wherever possible,
shrinking the judge's role to rationale checks and free-text answers. Where
an LLM judge is unavoidable, it ships with a \emph{decoy calibration set} ---
plausible-but-wrong answers a sound judge must reject at $\geq 95\%$ --- and
every published run reports its judge's decoy rejection rate next to its
scores. Decoys are generated per item from four templated wrong-answer
classes targeting the specific leniency failures LoCoMo's audit documented
(construction protocol: Appendix~\ref{app:prompts}); the class-wise
rejection breakdown is
reported, not just the aggregate, because a judge lenient to exactly one
class (the stale answer asserted as current) inflates exactly the metric
Track~C exists to expose. Measured per-judge rejection rates ship with
the Track~C instantiation in v2; Track~B needs no LLM judge (verdict and
attribution matching are structural).

\subsection{Separability audit}
Before release, three reference baselines run on every item: flat RAG,
fact-extraction RAG, and an oracle given gold evidence. Items where flat RAG
succeeds on Tracks A/B without contradiction-specific machinery, or where
the oracle fails, are rejected or rewritten. Pre-registered targets: flat
RAG $\leq 15\%$ detection F1 on Track A; oracle $\geq 95\%$. For Track~B
--- where flat RAG is \emph{expected} to detect blatant contradictions
--- the audit's role is played by the per-backend flat-RAG columns of
Table~\ref{tab:v1prelim} together with human validation (\S 5.1): what
separates systems is not whether contradictions can be found but whether
finding them costs false flags and lost attribution. The oracle half of
the audit \emph{is} run on Track~B as the gold-evidence condition, and
passes: given the cited gold turns, all three backends reach 1.000
contradiction recall (target $\geq$0.95), so no kept item is unsolvable
from its own evidence --- a detectability oracle; it bounds neither
specificity nor full-record reasoning. The formal three-baseline audit runs per item for Tracks A
and C before their v2 release.

\subsection{Reproducibility standards}
Pinned judge model and prompts, published harness, fixed retrieval budgets
per mode, one command per track, and --- for vendor self-reports --- a
linked raw per-item output file. We hold ourselves to the same standard:
every artifact behind this paper --- the frozen v1.0 key, candidate items
and generation rejects, both annotators' raw answers, adjudication queue
and decisions, council pre-screen, per-item outputs for all thirteen
configurations, and the statistics --- is at
\url{https://github.com/subratpanda/twist-benchmark}.

\section{Scoring and reporting}
\label{sec:scoring}
Per-track scores only; no single headline number (a composite invites gaming
and hides trade-offs). The canonical report is a five-column profile:
LoCoMo-J $\cdot$ A (detection F1~/~false-flag) $\cdot$ B
(contradiction recall~/~hard-negative specificity) $\cdot$ C
(current~/~stale) $\cdot$ D (leak~/~over-block). Paired metrics must be reported together --- citing
detection without false-flag rate is not a valid TWIST citation, a rule
stated in the usage terms. Online and batch modes are separate leaderboard
columns.

\section{Experiments}
\label{sec:experiments}

\subsection{Pre-registered predictions}
\label{sec:predictions}
To make the benchmark's discriminative claim falsifiable, the v0.1 spec
pre-registered expectations before any system ran: flat RAG scores near
floor on Track~A (no emission mechanism), moderately on Track~B ---
catching blatant contradictions while failing hard negatives and
attribution --- and poorly on Track~C stale-rate; systems with
contradiction machinery should separate from flat RAG; and if the
separations do not materialize, the benchmark has failed its purpose.
Verdict on the Track~B predictions against v1.0
(Table~\ref{tab:v1prelim}): \emph{catches blatant contradictions} ---
confirmed for every backend (flat RAG 0.76--0.97); \emph{fails hard
negatives} --- confirmed for GPT-4o (0.57 with retrieval, 0.37 with full
context), and partially for Claude/Gemini (0.74--0.84 with retrieval):
better-calibrated backends narrow but do not close the governance gap
against the coherence-oriented system's 0.98--1.00;
\emph{attribution is hard for retrieval-bound systems} --- confirmed
(0.44--0.59 for everything that must retrieve, vs.\ 0.97--1.00 when
gold evidence is given: an evidence-grounding limit, not a reasoning
limit). \emph{Separation} --- confirmed on the specificity and
attribution columns, refuted on raw detection recall, where flat RAG
dominates. We claim only that the tested systems occupy different
regions of the recall/specificity/attribution trade-off, not that
architecture is the isolated cause: reader model and memory architecture
are confounded across most of Table~\ref{tab:v1prelim}. The exception is
the model-controlled GPT-4o triple --- the deployed system's internal
LLM, the flat-RAG baseline, and the full-context baseline all use GPT-4o
--- where one model under three memory regimes spans opposite corners
(0.42/0.98, 0.97/0.57, and 0.95/0.37 recall/specificity), the clearest
model-held-constant contrast in the data (prompt shape, call structure,
and evidence format still differ across regimes, so even this is not
pure architectural causality). A fully controlled
architecture sweep (one model, one prompt, one retrieval budget, memory
mechanism varied) is the natural v2 experiment.

\subsection{Pilot: Track B v0}
\label{sec:pilot}
A 153-item Track~B pilot (93 contradicting / 41 aligned / 19 hard-negative
drafts over the 10 LoCoMo conversations) produced the benchmark's central
early evidence: \textbf{a flat-RAG baseline and a deployed
coherence-oriented system fail in opposite directions}
(Table~\ref{tab:pilot}). Flat RAG flags every true contradiction but also
63\% of hard negatives --- deployed, an assistant that objects to safe
messages two times out of three. The deployed system never over-flags
(1.000 on both governance columns) but catches only 11--14\% of true
contradictions at 10-conversation scale. A recall-only benchmark sees
neither failure. The pilot also surfaced the role-flip item defect that
became item-quality rule~1 (\S\ref{sec:item-rules}): in most residual
misses, the gold evidence turn's standalone text did not carry the
contradicted claim.

\begin{table}[ht]
\centering\small
\caption{Track B v0 pilot (GPT-4o backend; superseded by the validated
v1.0 results in Table~\ref{tab:v1prelim} --- shown for the diagnostic
history: v0 items were not human-annotated and its hard-negative subset
was small, $n{=}19$). Paired metrics; neither column is citable alone.}
\label{tab:pilot}
\begin{tabular}{lcc}
\toprule
Metric & MindTwin & Flat RAG \\
\midrule
Contradiction recall & 0.108 & \textbf{1.000} \\
Aligned specificity & \textbf{1.000} & 0.780 \\
Hard-negative specificity & \textbf{1.000} & 0.368 \\
Attribution accuracy & 0.500 & 0.419 \\
\bottomrule
\end{tabular}
\end{table}

\subsection{v1.0 results}
\label{sec:v1-prelim}
Table~\ref{tab:v1prelim} reports thirteen configurations on the frozen,
human-validated v1.0 key (161 items; \S\ref{sec:validation}): a
\emph{draft-only} floor (the draft and nothing else --- any accuracy
above always-consistent quantifies label leakage through draft style),
the deployed system, flat RAG, a \emph{full-context} baseline (the
entire dated transcript in the prompt --- isolating contradiction
reasoning from retrieval), and the \emph{gold-evidence condition} (the
cited gold turns only --- an oracle for contradiction
\emph{detectability} and evidence sufficiency, not a global upper
bound: restraint degrades on bare excerpts), each across three LLM
backends. Wilson 95\% intervals accompany the small-$n$ columns;
conversation-clustered bootstrap intervals and exact pairwise McNemar
tests with Holm correction are released with the results, and for the
stochastic vault-based rows the primary uncertainty statement is the
multi-ingest analysis below. Six observations:
(1)~\textbf{Gold-evidence recall is 1.000 for all three models}: every
gold contradiction is detectable from its evidence alone, meeting the
pre-registered oracle target ($\geq$0.95) and independently
corroborating the human-validated key.
(2)~\textbf{For calibrated models, the remaining failure is retrieval,
not reasoning.} Given the full transcript, Claude reaches
0.947~/~0.951~/~0.919 with 0.889 attribution --- close to solving the
track --- and Gemini is similar; their flat-RAG deficits (e.g., Gemini recall 0.76 vs.\ 0.92
full-context) are consistent with substantial retrieval-coverage
limitations (full context also changes chronology and disambiguation
cues, so we do not claim retrieval is the sole cause). GPT-4o is the counterexample: its hard-negative specificity is
poor with retrieval (0.57), \emph{worse} with the full transcript
(0.37), and still poor with gold evidence only (0.42) --- its
over-flagging is model-intrinsic, not context-induced.
(3)~\textbf{How context shapes restraint is strongly model-dependent.}
Claude shows a steep context gradient --- 0.50 hard-negative specificity
given bare gold excerpts, 0.92 with the full transcript --- consistent
with contrastive framing looking damning without surrounding context;
Gemini is comparatively stable across all three conditions
(0.84--0.90); and GPT-4o is over-sensitive everywhere, with its
\emph{worst} specificity at full context (0.37). Context poverty is one
mechanism behind retrieval baselines crying wolf, but it is not
universal --- which condition harms restraint most depends on the model.
(4)~\textbf{Draft-only floors are strongly model-dependent --- and
GPT-4o's is high.} With no record at all, GPT-4o flags 71\% of
contradicting drafts at 0.85/0.90 specificity; Claude reaches 0.34 and
Gemini 0.03. Contradicting drafts evidently carry a recognizable
negative-presumptive style that some models exploit as a prior. Every
record-grounded configuration significantly exceeds its own draft-only
floor (Holm-adjusted McNemar $p<0.05$), and grounding is indispensable
for evidence: draft-only attribution is zero by construction. Still,
this is measurable leakage --- v1.1 will add style-matched aligned
drafts (negative-presumptive drafts that happen to be \emph{true}) to
price style priors the way hard negatives price topic overlap.
(5)~\textbf{Grounded contradiction recall separates intervening from
intervening-and-knowing-why.} Unconditional GCR (correct flag with
valid top-3 evidence, over all 38 items) orders the ladder: deployed
0.18, flat RAG 0.45--0.55, full-context Claude 0.84, gold-evidence
0.97--1.00 --- conditional attribution alone would hide that the
deployed system's strong-looking 0.44 is computed over only 16 flags.
(6)~\textbf{No tested configuration simultaneously achieves high
contradiction recall, high hard-negative specificity, and high
grounded recall.} The deployed system is the extreme restraint point
--- 1.000~/~0.984 specificity, recall 0.421 [0.28, 0.58],
significantly below every record-grounded baseline (exact McNemar,
Holm-adjusted $p<0.05$ pairwise) --- while the best full-context
configuration still concedes 8\% false flags on hard negatives. The over-flagging--versus--silence trade-off measured
here at the memory-system layer mirrors the
drift-resistance--versus--update-tracking trade-off BeliefShift reports
at the model layer \citep{myakala2026beliefshift}: two benchmarks, two
layers, the same stability--plasticity tension.

Two honesty notes. Vault-based rows (deployed system, flat RAG) vary
across re-ingestions --- fact extraction and index construction are
stochastic --- so for those rows the primary uncertainty statement
pools three independent ingests of the v1.0 key (two with per-item
outputs, one score-level from a partially completed run; all released):
deployed-system recall 0.32/0.42/0.42, multi-ingest mean 0.39 with
conversation-$\times$-ingest hierarchical bootstrap 95\% interval
[0.19, 0.51]; GPT-4o flat-RAG hard-negative specificity is itself
ingest-sensitive (0.42--0.57, mean 0.48 [0.37, 0.66]). Interval
estimates, not point deltas, are the citable quantity (the 0.90 vs.\
0.91 flat-RAG backend differences are noise), and attribution/GCR
variance across ingests is estimable only from the cap-consistent
ingest --- a stated limitation; a three-fresh-ingest protocol is the
venue upgrade. And item validation lifted every system's
detection score relative to the pre-annotation candidates (e.g., Gemini
flat RAG 0.65$\to$0.76--0.84): the adjudicated drops were items systems
could not reasonably flag, corroborating that annotation removed genuine
defects rather than inconvenient results.

\begin{table}[ht]
\centering\small
\caption{Track B on the frozen \textbf{TWIST-v1.0} key ($n{=}161$;
retained-set $\kappa=0.851$). CR = contradiction recall ($n{=}38$),
AS = aligned specificity ($n{=}61$), HNS = hard-negative specificity
($n{=}62$), GCR = grounded contradiction recall (correct flag with
valid top-3 evidence, over all 38), Attr = conditional attribution over
a top-3 budget. Brackets: Wilson 95\%. Vault-based rows ($\dagger$)
are stochastic across ingests; their multi-ingest means and
hierarchical intervals are in the text. Paired metrics: no column is
citable alone; per-item outputs, all intervals, and Holm-adjusted
McNemar tests are released with the benchmark.}
\label{tab:v1prelim}
\setlength{\tabcolsep}{3.5pt}
\begin{tabular}{llccccc}
\toprule
System & Backend & CR & AS & HNS & GCR & Attr \\
\midrule
Draft only (no record) & GPT-4o & 0.71 [.55,.83] & 0.85 & 0.90 & 0.00 & --- \\
Draft only (no record) & Claude & 0.34 [.21,.50] & 0.97 & 0.95 & 0.00 & --- \\
Draft only (no record) & Gemini & 0.03 [.01,.14] & 1.00 & 1.00 & 0.00 & --- \\
\midrule
Deployed (MindTwin)$^\dagger$ & GPT-4o & 0.42 [.28,.58] & \textbf{1.00} & \textbf{0.98} [.91,1.0] & 0.18 & 0.44 \\
\midrule
Flat RAG$^\dagger$ & GPT-4o & \textbf{0.97} [.87,1.0] & 0.77 & 0.57 [.44,.68] & 0.55 & 0.57 \\
Flat RAG$^\dagger$ & Claude & 0.92 [.79,.97] & 0.90 & 0.74 [.62,.83] & 0.50 & 0.54 \\
Flat RAG$^\dagger$ & Gemini & 0.76 [.61,.87] & 0.97 & 0.84 [.73,.91] & 0.45 & 0.59 \\
\midrule
Full context & GPT-4o & 0.95 [.83,.99] & 0.74 & 0.37 [.26,.50] & 0.66 & 0.69 \\
Full context & Claude & 0.95 [.83,.99] & 0.95 & \textbf{0.92} [.83,.97] & \textbf{0.84} & \textbf{0.89} \\
Full context & Gemini & 0.92 [.79,.97] & \textbf{0.97} & 0.90 [.81,.96] & 0.63 & 0.69 \\
\midrule
Gold evidence & GPT-4o & 1.00 [.91,1.0] & 0.75 & 0.42 [.31,.54] & 0.97 & 0.97 \\
Gold evidence & Claude & 1.00 [.91,1.0] & 0.82 & 0.50 [.38,.62] & 1.00 & 1.00 \\
Gold evidence & Gemini & 1.00 [.91,1.0] & 0.93 & 0.87 [.77,.93] & 1.00 & 1.00 \\
\bottomrule
\end{tabular}
\end{table}

\paragraph{History-length stratification (rule 5).} Splitting v1.0 at
620 history turns (74 items over $<$620-turn conversations, 87 over
longer ones): the deployed system's contradiction recall falls from
0.48 to 0.31 as history grows --- the retrieval-coverage stress Track~B
is built to apply --- and flat-RAG recall and restraint both degrade
(Gemini recall 0.84$\to$0.62; Claude hard-negative specificity
0.78$\to$0.71), while the full-context configurations are essentially
length-robust (Claude 0.92$\to$1.00 recall). Longer records make
silence harder for retrieval-bound systems and restraint harder for
retrieval baselines; strata are small ($n{=}13$ long-history
contradicting items), so we report direction, not significance.


\section{Limitations and ethics}
\label{sec:ethics}
\textbf{Synthetic data.} Items are LLM-drafted; naturalness is bounded by
generation quality. Mitigations: persona-matched generation against real
LoCoMo turns, structural gates, human annotation, and the CS track's
structural (not merely stylistic) realism. Real-workspace validation
remains future work with consenting design partners.
\textbf{Single-generator circularity.} Track B v1 items were generated and
gate-verified by one model family (GPT-4o); a system built on the same
family may share blind spots with the item distribution. Disclosed
mitigations: the structural gates are model-free, the council pre-screen
adds two independent model families, and human annotation is the gold
gate. \textbf{Crisis content (Track D).} Crisis plants use established,
non-graphic test phrasings, ship in a separated opt-in file with content
warnings, and follow published crisis-line guidance; the goal is measuring
\emph{governance}, not building a crisis-classifier corpus.
\textbf{Author bias.} TWIST is proposed by a vendor whose product is built
around these capabilities. The mitigations are structural: public gold
keys and per-item outputs, a public harness that itself runs the reference
baselines, the separability audit --- item rejection for tracks where
plain retrieval should not suffice, and a gold-evidence detectability
check for Track~B --- an errata process, and the requirement that
$N\geq2$ external
systems be run by their own authors before ``benchmark'' is used without
``proposed'' in front of it. A datasheet is provided in
Appendix~A \citep{gebru2021datasheets}.

\section{Conclusion}
\label{sec:conclusion}
Retrieval correctness is necessary but insufficient for persistent
memory. A deployable memory system must intervene correctly at belief
change points --- notice unresolved tensions unprompted, stop a draft
that contradicts the record, answer with the current belief while
keeping its history, govern sensitive recall --- and it must know when
not to intervene, which is why every TWIST metric ships with a matched
false-intervention control. TWIST evaluates that intervention boundary
at the memory-system layer. Track B is instantiated
here with a validation pipeline whose structural gates, documented
generator-failure taxonomy, and disclosed model-council pre-screen are
themselves reusable methodology for anyone building contradiction
benchmarks. Tracks A and C (injection pipelines specified in
\S\ref{sec:tracks}) and the CS-domain corpus are v2; Track D ships with
its opt-in safety corpus and clinical-review process. The benchmark
becomes credible exactly when competitors run it themselves and dispute
items through the errata process --- both are design requirements, and we
invite the authors of the systems and benchmarks cited here --- via the
public repository above --- to be the
first external runs.

\section{Acknowedgment}
\label{sec:ack}
The author would like to thank Mr. Marvin Danig (From Achiral.ai) for the insightful discussions that led to this work and Mr. Kinshuk Attri for helping with annotations.

\section{Declaration}
\label{sec:decl}
Declaration of generative AI and AI-assisted technologies in the manuscript preparation process

During the preparation of this work, the author(s) used Claude for text generation, summarization and critical analysis. The author reviewed and edited the output as needed and take full responsibility for the content of the published article.

\bibliographystyle{plainnat}
\bibliography{references}

\appendix
\section{Datasheet for TWIST (Track B v1)}
Condensed per \citet{gebru2021datasheets}; full version with the release.
\textbf{Motivation.} Created to make coherence/alignment/governance claims
about conversational memory systems falsifiable; funded and built by
MindTwin (author bias mitigations: \S\ref{sec:ethics}).
\textbf{Composition.} 200 draft-vetting items over the 10 public LoCoMo
conversations: 74 contradicting (gold \texttt{aligned=false} + evidence
turn ids + verbatim contradicted quote), 62 aligned, 64 hard-negative
(both gold \texttt{aligned=true}); per-item scenario family and
history-length metadata. No real persons; LoCoMo personas are synthetic.
\textbf{Collection.} LLM-generated from parameterized family templates
against LoCoMo transcripts; three structural gates plus a
verification-as-seen LLM gate (\S\ref{sec:pipeline}); $\sim$490 generated,
200 kept; all rejects published with verdicts.
\textbf{Labeling.} Gold from generation, pre-screened by a three-model
council, validated by two independent gold-blind human annotators with
author-led adjudication (disclosed) ($\kappa$ and drop counts: \S\ref{sec:validation}).
\textbf{Uses.} Evaluating draft-alignment vetting in memory systems; not a
training set --- the held sets are frozen and tuning against them is
prohibited by the usage terms.
\textbf{Distribution.} At
\url{https://github.com/subratpanda/twist-benchmark}: items, gold,
harness, annotation workbooks, council
outputs, and per-item system outputs at the public repository; LoCoMo
corpus under its original license.
\textbf{Maintenance.} Versioned keys (v1.0, v1.1, \dots) with a public
errata process; scores must cite the key version.
\section{Scenario family catalog}
Each detection family is paired with a surface-matched hard-negative
family; an item instantiates one family against one conversation with
machine-readable gold. Track B examples are real v1 items.

\begin{table}[ht]
\centering\small
\caption{Track B families (instantiated). HN families carry gold
\texttt{aligned=true}.}
\begin{tabular}{p{3.4cm}p{8.6cm}}
\toprule
B1 invented activity & Draft congratulates the recipient on something the
record positively conflicts with: \emph{``all set for the big marathon next
week!''} vs.\ ``Messed up my knee playing b-ball.'' \\
B2 negated commitment & Draft assumes the recipient dropped an active
commitment: \emph{``enjoying your break from adoption plans''} vs.\ ``I
passed the adoption agency interviews last Friday!'' \\
B3 stale assumption & Draft locks in a state the record shows has changed:
\emph{``how is the banking project progressing?''} vs.\ ``Lost my job as a
banker yesterday.'' \\
B-AL consistent & Draft matches the record's current state. \\
HN-B1 committed-despite-challenges & Contrastive framing over an affirmed
commitment (``awesome you're \emph{still} committed\dots'') --- bait, not
contradiction. \\
HN-B2 acknowledged pause & Draft acknowledges a break the record contains,
rather than misstating it. \\
HN-B3 change-question & Draft \emph{asks} whether something changed;
asking is never asserting. \\
\bottomrule
\end{tabular}
\end{table}

Track A families (specified): A1 direct stance reversal, A2 drift-then-flip,
A3 cross-channel reversal (CS track); hard negatives: nostalgia,
acknowledged change of mind, hypothetical/joke, pressure-signal-alone.
Track C: single update, double update, history-aware query,
post-resolution; hard negative: coexisting facts. Track D: direct PII
probe, indirect aggregation probe, crisis plant, referral correctness;
hard negative: emotional-but-not-crisis content that must remain
recallable. Full templates ship with the repository.

\section{Prompts and judge decoy protocol}
\label{app:prompts}
The verification-as-seen gate prompt (abridged; verbatim prompts, including
the generation and council-annotator prompts, ship in the repository):
\begin{quote}\small
``You are validating an item for a benchmark of memory systems. You see
ONLY (a) a few excerpts retrieved from a long conversation record, exactly
as a vetting system would see them, and (b) a draft message someone
proposes to send to a named person. Judge strictly from the excerpt text
[\dots] 1.~\texttt{contradicts} --- would sending the draft contradict the
record? The dated excerpts are the LATEST known record --- do not excuse an
incompatibility by imagining undocumented later changes. Acknowledging a
change, asking a question about a possible change, or merely discussing
the same topic is NOT a contradiction. 2.~\texttt{claim\_stated} --- do
the excerpts BY THEMSELVES explicitly state, in the named person's own
words, the fact or stance at issue? Answer false if identifying it
requires guessing who a plan belongs to, resolving a pronoun, or reading
between the lines of a reply.''
\end{quote}

\paragraph{Judge decoy set construction (Track C).}
For every free-text probe (current-state and history-aware queries), the
harness generates one decoy per wrong-answer class and mixes decoys with
gold answers in randomized order; the judge grades the mixed set blind to
provenance. The four classes are templated from the update chain's own
structure, so decoys are plausible by construction rather than by prompt
luck:
\begin{enumerate}[leftmargin=1.6em,itemsep=1pt,topsep=2pt]
  \item \textbf{Stale-as-current}: the superseded belief $F_0$ (or $F_1$ in
    a two-step chain) asserted fluently as the current answer. This is the
    load-bearing class --- it is the exact failure Track~C measures, so a
    judge that accepts it corrupts the stale-answer rate directly.
  \item \textbf{Perturbed detail}: the correct current fact with one
    critical slot altered (date, entity, location, or direction of change),
    testing whether the judge verifies content or rewards topical fluency.
  \item \textbf{Fluent non-answer}: a well-formed response that restates
    the question, hedges, and commits to no belief state --- the
    politeness-leniency failure documented in LoCoMo's audit.
  \item \textbf{Unordered both-states} (history-aware queries only): both
    $F_0$ and $F_1$ asserted without temporal ordering. History-aware gold
    requires supersession \emph{order}; a judge accepting the unordered
    form collapses the track's distinction between retention and revision.
\end{enumerate}
Rejection rate is computed per class and per judge backend; a judge is
admissible for a published run only if every class clears the $\geq 95\%$
bar. Decoy templates and generation prompts ship in the repository; the
decoy items themselves are published so judge calibration is independently
re-runnable. Measured per-class rejection rates for GPT-4o, Claude, and
Gemini judges are reported with the Track~C instantiation (v2); no result
in this paper depends on an LLM judge.

\section{Crisis-content handling}
\label{app:crisis}
Track~D's crisis plants are governed by three mechanisms, stated here as
binding release requirements for the v2 corpus.

\paragraph{Opt-in distribution.} Crisis plants ship in a separate file that
is excluded from the default dataset download and clone path. Obtaining it
requires an explicit harness flag (\texttt{--include-crisis}) that prints a
content notice and requires interactive confirmation (or a documented
non-interactive acknowledgment variable for CI). Every other component of
TWIST --- all of Tracks A--C and Track~D's PII and emotional-hard-negative
items --- runs without the file; a run without it reports Track~D as
``PII-only'' rather than a full Track~D score, so no one is pressured into
downloading crisis content to appear on the leaderboard. The file carries a
content-warning header, and every item is tagged with its scenario family
so downstream tooling can filter mechanically.

\paragraph{Guidance sources.} Plants use established, non-graphic test
phrasings: indirect ideation and disclosure patterns of the kind published
in safe-messaging guidance for media and crisis services (WHO
suicide-reporting guidance and national crisis-line safe-messaging
standards), never method details, plans, or graphic content. All personas
are synthetic; no real person's crisis is represented. Referral-correctness
gold is region-aware --- the correct behavior maps to the crisis resource
of the conversation's stated locale --- because a memory system deployed
across regions that emits a single hard-coded hotline fails real users.
The specific guidance documents consulted are enumerated in the corpus
release notes so the construction basis is auditable.

\paragraph{Annotator safety.} Crisis items are segregated into a separately
labeled workbook block that an annotator may decline in full without
affecting the validity of their pass on all other items (the annotation
protocol treats the crisis block as independently adjudicable). The block
is preceded by a content notice, annotators see the smallest sufficient
excerpt rather than full conversations, no time-pressure element of the
protocol applies to it, and adjudication of crisis items never requires
re-reading surrounding sessions. Crisis-line contact information for the
annotator's own region is included in the workbook header.

\end{document}